\documentclass[journal]{IEEEtran}
\ifCLASSINFOpdf
\else
\fi
\usepackage{amsmath}
\usepackage{graphics}
\usepackage{graphicx}
\usepackage{epstopdf}
\usepackage{amssymb}
\usepackage{algorithm,algorithmic}
\usepackage{hyperref}
\usepackage{url}
\usepackage{multirow}
\usepackage{array,ragged2e}
\newcolumntype{M}[1]{>{\centering\arraybackslash}m{#1}}
\usepackage{float}
\usepackage{cite}
\usepackage[labelfont=bf]{caption}
\usepackage{caption}
\usepackage{subcaption}
\usepackage{xcolor}

\begin{document}
%
\title{A Contactless Fingerprint Recognition System}
%
%
%

\author{Aman~Attrish,
        Nagasai~Bharat,
        Vijay~Anand,~\IEEEmembership{Member,~IEEE,}
        and Vivek~Kanhangad,~\IEEEmembership{Senior Member,~IEEE}
\thanks{A. Aman, N. Bharat, V. Anand and V. Kanhangad are with Discipline
of Electrical Engineering, Indian Institute of Technology Indore, Indore 453552, India. e-mail:(ee160002003@iiti.ac.in, ee160002016.iiti.ac.in, vijayanand.phd20@gmail.com, kvivek@iiti.ac.in)}}

%
%

\markboth{}%
{Shell \MakeLowercase{\textit{et al.}}: Development of a Contact-less Biometric System}
%



\maketitle

\begin{abstract}
 Fingerprints are one of the most widely explored biometric traits. Specifically, contact-based fingerprint recognition systems reign supreme due to their robustness, portability and the extensive research work done in the field. However, these systems suffer from issues such as hygiene,  sensor degradation due to constant physical contact, and latent fingerprint threats. In this paper, we propose an approach for developing a contactless fingerprint recognition system that captures finger photo from a distance using an image sensor in a suitable environment. The captured finger photos are then processed further to obtain global and local (minutiae-based) features. Specifically,  a siamese convolutional neural network (CNN) is designed to extract global features from a given finger photo. 
The proposed system computes matching scores from CNN-based features and minutiae-based features.  Finally, the two scores are fused to obtain the final matching score between the probe and reference fingerprint templates. Most importantly, the proposed system is developed using the Nvidia Jetson Nano development kit, which allows us to perform contactless fingerprint recognition in real-time with minimum latency and acceptable matching accuracy. The performance of the proposed system is evaluated on an in-house IITI contactless fingerprint dataset (IITI-CFD) containing 105 train and 100 test subjects. The proposed system achieves an equal-error-rate of 2.19\% on IITI-CFD.
\end{abstract}


%
\IEEEpeerreviewmaketitle

\section{Introduction}
%
%
%
%
\IEEEPARstart{F}{aced} with challenges of identity theft from password and PIN-based authentication systems, new technological solutions are gradually being applied. One of these technologies, biometrics, has quickly proved itself as the most suitable means of authenticating and identifying individuals in a fast, safe, and reliable way, using unique biometric traits.

In overview,  fingerprints are the most widely explored biometric traits. A fingerprint image consists of patterns of ridges and valleys (furrows) found on the fingertip \cite{Moenssens}. The automated fingerprint recognition systems (AFRS) recognize a person by a traditional method of matching a fingerprint pattern. The fingerprint recognition can be carried out in two ways, namely; fingerprint verification and fingerprint identification \cite{maltoni2009handbook}. In the verification process, the biometric system performs a one-to-one matching of the user's fingerprint with the fingerprint template stored in the database to verify if the claimed identity is true or false \cite{jainonline}. On the other hand, in the identification process, the user's fingerprint is compared to all the fingerprint templates stored in the database to obtain the user's identity. Hence, the identification process is computationally expensive as compared to the verification process, especially for large databases \cite{Moenssens}.

Generally,  in contact-based AFRS, the fingerprint image is captured with the incorporation of an advanced complementary MOSFET (CMOS) image sensor\cite{maltoni2009handbook}. Non-linear spatial distortions and low contrast regions due to improper pressures of the finger on the sensor platen are some of the challenges that are common in contact-based biometric systems \cite{maltoni2009handbook}. 
Due to the constant physical contact of each individual's finger with the sensor in contact-based AFRS, there might be an issue where the cleanliness of the sensor should be of the most importance. This can lead to the spread of contagious diseases to users; also, the system might not work as expected due to the accumulation of  dust and dirt on the sensor. Further, contact-based sensors have a high maintenance cost as they can be faulted easily when not appropriately used during physical contact. Furthermore, these technologies encounter a significant security threat, since every acquisition of fingerprint leaves a latent print of the finger on the sensor surface, which could be easily lifted off the sensor surface. 

The solution to the aforementioned issues leads to the development of the biometric system in the contactless domain using a camera sensor, which captures the fingerprint image in a suitable capturing environment. 
Piuri and Scotti \cite{4699379} 
 investigated techniques to suitably processing the camera images of fingertips such that the processed images are similar to the fingerprint images captured using the dedicated sensor. The primary focus of the  work presented in \cite{4699379} is to leverage the existing contact-based fingerprint recognition techniques to develop a contactless fingerprint recognition system using fingerprint images from mobile camera and webcam.
Labati \emph{et al.}\cite{6607909} proposed an approach to recover perspective deformations and improper fingertip alignments in single camera systems in the contactless domain of the biometric finger systems. This approach incorporated Sony CCD camera sensor. It is done to eliminate the non-idealities of the contactless acquisition of fingerprint samples. To improve the visibility of the ridge patterns, illumination techniques presented in \cite{1663788,4448572,4150099} are used.
Lin and Kumar\cite{8409476} presented a CNN-based framework to  match contactless and contact-based fingerprint images.
Michael \emph{et al.} \cite{article} developed a biometric system using visible and infrared imagery on the five features namely hand geometry, palm print, palmar knuckle print, palm vein, and finger vein for the recognition. Kumar \cite{8768216} investigated the possibility of recognizing completely contactless finger knuckle images acquired under varying poses. Experimental results presented in \cite{8768216} validate the usefulness of normalization and matching algorithms to recognize finger knuckle with different poses.

A review of the literature indicates that contactless fingerprint biometrics has not been explored much, and most of the existing research has been focused only on the analysis and simulation aspects of contactless fingerprint biometrics. Despite the advances in sensor technology and edge computation power, very little research work has implemented the algorithms on hardware and prototype a biometric system. This motivated us to develop a contactless fingerprint biometric system considering the significant research in the field of biometrics and deep learning and implement it in a system to be used in real-time. 

The objective of our work is to develop a contactless fingerprint recognition system (CFRS), incorporating both deep learning and standard fingerprint matching algorithms.  The primary focus is on the implementation of CFRS in real-time on hardware setup with minimum latency and high matching accuracy.
Major contributions of this work are as follows: 
a customized siamese CNN architecture has been designed  in accordance with the finger images captured from the camera sensor in the system.  
The siamese CNN network along with minutiae-based matching algorithm have been deployed on Nvidia Jetson Nano kit to 
develop a CFRS  in real-time with minimum latency and acceptable matching accuracy. 
 

{The rest of the paper is organised as follows: the proposed approach is described in Section II, in which a brief overview of the challenges faced in the contactless domain, the proposed algorithm and the incorporated NIST software are presented. This is followed by the detailed description of the methods employed for the problems faced using deep learning and state-of-the-art techniques. The hardware incorporated in the project are well-demonstrated next. Experimental results and discussion are presented in Section III. This section also presents a brief description of the database developed and the performance measures used for evaluating the performance of the system. Finally, conclusion and future work are presented in Section IV.}

\section{Proposed Approach}
We have developed a CFRS consisting of three major components namely, contactless finger image capturing module, CNN based global feature  matching module and minutiae  feature matching module.
 The proposed CFRS captures finger image from a distance using  Raspberry Pi NoIR camera V2 that has a Sony IMX219 8-megapixel sensor. 
 \begin{figure*}[ht!]
    \centering
    \includegraphics[width = 0.9\textwidth]{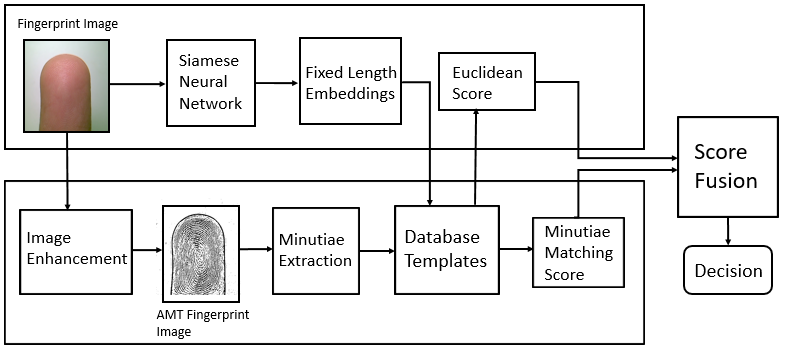}
    \caption{Schematic diagram of proposed parallel approach}
    \label{parallel_approach}
\end{figure*}
A schematic diagram of the proposed approach is presented in Fig. \ref{parallel_approach}. As can be observed, the proposed approach employs a customized siamese CNN architecture to process images captured from the camera sensor in the system. Specifically,  siamese network generates a fixed length embedding of the fingerprint image which is then utilized to calculate a similarity score between the probe and the reference images.
Further, we have employed image enhancement technique on the captured finger image and then  performed minutia-based matching using the standard  NIST Biometric Image Software (NBIS). Finally,  the scores obtained from both of the modules are fused to obtain the final score.

\begin{figure*}[t]
    \centering
    \begin{subfigure}[b]{0.3\textwidth}
        \centering
        \includegraphics[width=\textwidth]{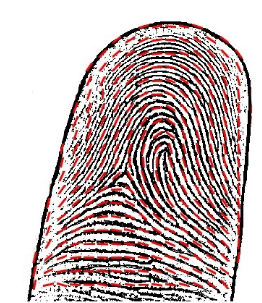}
        \caption{Orientation map of a fingerprint}
        \label{fig:orientation_map}
    \end{subfigure}
    \hfill
    \begin{subfigure}[b]{0.3\textwidth}
        \centering
        \includegraphics[width=\textwidth]{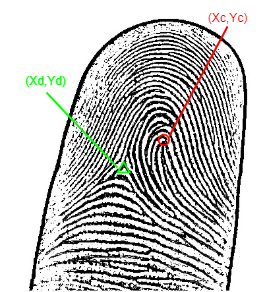}
        \caption{Core and delta points location}
        \label{fig:singular_points}
    \end{subfigure}
    \hfill
    \begin{subfigure}[b]{0.3\textwidth}
        \centering
        \includegraphics[width=\textwidth]{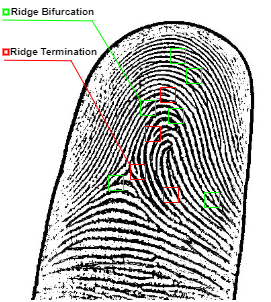}
        \caption{Types of minutiae of a fingerprint}
        \label{fig:minutiae_type}
    \end{subfigure}
    \caption{Global and local features extracted from fingerprint image}
    \label{features}
\end{figure*}

We have divided our approach in two stages \textit{i.e.}, developing algorithms and modifying the existing ones to increase the matching accuracy of two fingerprint templates, and implementing it on hardware with minimum latency. Generally, it is challenging to deal with contactless fingerprint images due to problems like perspective distortion and deformation as discussed in \cite{6607909}. Lightning conditions affect the quality of the image, and the amount of information of fingerprint captured by the image sensor.
To circumvent the aforementioned issues, we have to consider all the global features (orientation map, core and delta point locations; Fig. \ref{fig:orientation_map} and \ref{fig:singular_points}) as well as the local features (minutiae information; Fig. \ref{fig:minutiae_type}) for extracting the maximum information from a fingerprint image captured by the image sensor. 
Thus, we have proposed a parallel approach (Fig. \ref{parallel_approach}) using deep learning to deal with global features \cite{ZHANG2019139} and state-of-the-art minutiae matching approach to deal with local features. Next, we provide a detailed description of each of the modules involved in the proposed CFRS.

\subsection{Minutiae-based approach}
\subsubsection{Ridge-valley map extraction}
\begin{figure*}[h]
    \centering
    \begin{subfigure}[b]{0.2\textwidth}
    \centering
    \includegraphics[width=\textwidth]{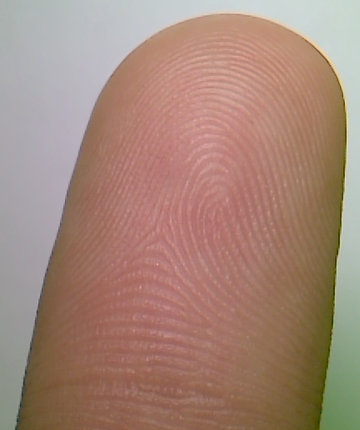}
    \caption{Raw image}
    \label{raw_image}
    \end{subfigure}
    \hfill
    \begin{subfigure}[b]{0.2\textwidth}
    \centering
    \includegraphics[width=\textwidth]{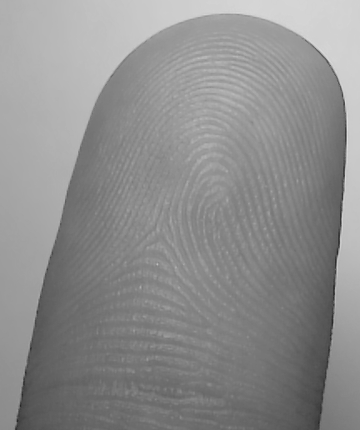}
    \caption{Gray scale image}
    \label{gray_image}
    \end{subfigure}
    \hfill
    \begin{subfigure}[b]{0.2\textwidth}
    \centering
    \includegraphics[width=\textwidth]{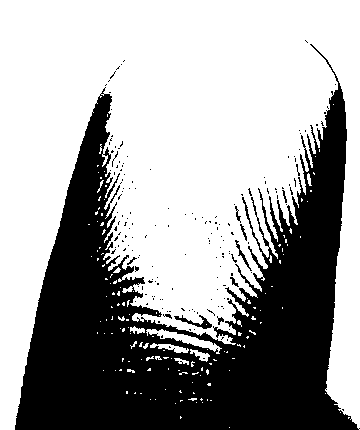}
    \caption{Global threshold}
    \label{global_thresh}
    \end{subfigure}
    \hfill
    \begin{subfigure}[b]{0.2\textwidth}
    \centering
    \includegraphics[width=\textwidth]{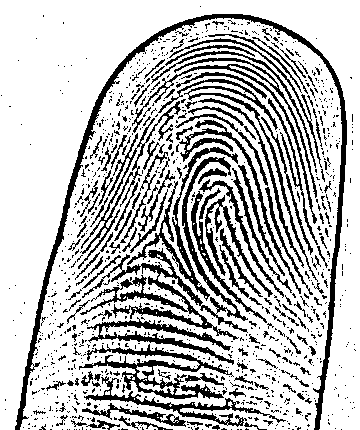}
    \caption{Adaptive mean threshold}
    \label{amt_image}
    \end{subfigure}
    \caption{Results from different thresholding techniques}
    \label{amt}
\end{figure*}

For the proposed CFRS, the image enhancement part plays a vital role since the information captured by the image sensor is low as compared to images from contact-based sensors. 
Firstly, there is a need to extract the ridge-valley map from the given fingerprint image. We have employed adaptive mean thresholding (AMT) on gray scale image to separate a foreground pattern of interest and the background image based on the difference in pixel intensities of each area \cite{Davies}.  AMT works on the principle  that the smaller image regions have roughly consistent illumination, hence more suitable for thresholding as compared to global thresholding. It can aid varying lighting states in the fingerprint image, e.g., those appearing as a result of a string glow, shadows, and gradients. Fig. \ref{amt} summarizes the result of the thresholding technique. First raw image (fig. \ref{raw_image}) is converted to gray-scale image (Fig. \ref{gray_image}) and then the ridge-valley map is extracted using AMT (Fig. \ref{amt_image}). The AMT technique clearly works better than the global thresholding (Fig. \ref{global_thresh}).

\subsubsection{Minutiae matching}
\begin{figure}[h]
    \centering
    \includegraphics[scale=0.35]{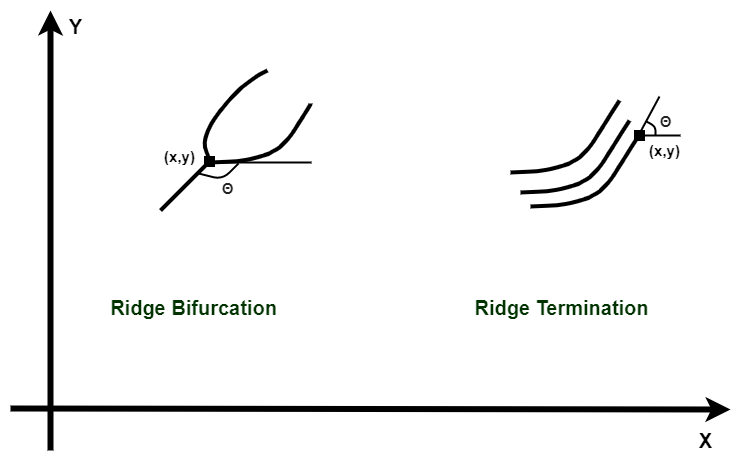}
    \caption{Minutia information}
    \label{minutiae}
\end{figure}

As presented in Fig. \ref{fig:minutiae_type}, local  features include ridge termination and ridge bifurcation, collectively called minutiae.
For minutiae extraction, we have used the standard  NBIS minutiae detector  MINDTCT \cite{NBIS}. It automatically detects and tracks records of minutiae information in the form of triplet
$[x, y,\theta]$, where $(x,y)$ is the position of minutia point and $\theta$ represents the orientation (Fig. \ref{minutiae}). The minutiae information  obtained by the MINDTCT algorithm is then used to perform matching.
Specifically, NBIS fingerprint matching algorithm, BOZORTH3, is incorporated for minutiae matching \cite{NBIS}. It is a minutia information based fingerprint matching algorithm that calculates similarity scores $(S_m)$ using matched minutiae. 

\subsection{Deep learning-based approach}
\begin{figure*}
    \centering
    \includegraphics[scale=0.25]{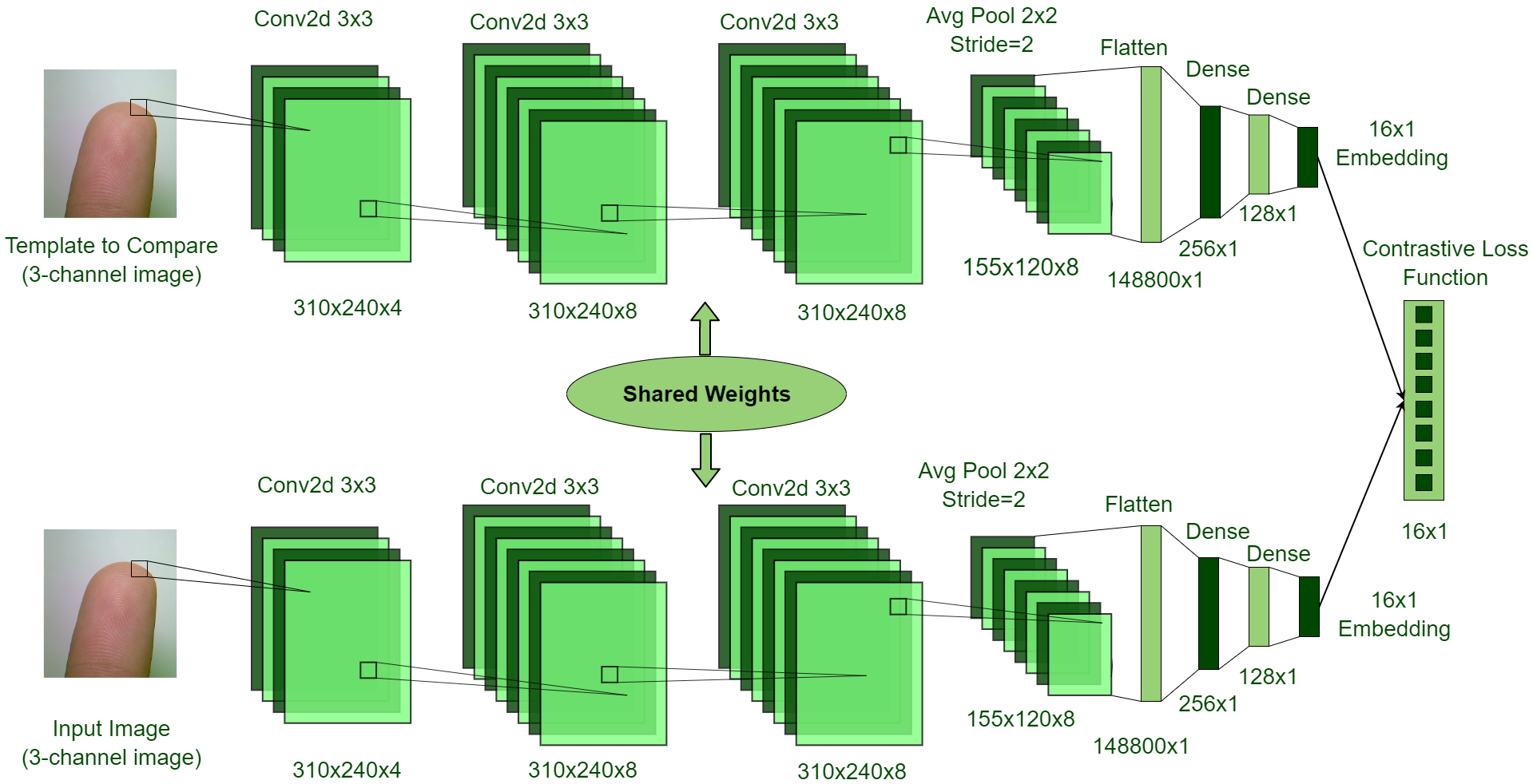}
    \caption{Architecture of siamese network}
    \label{siamese}
\end{figure*}

Recently, CNNs have been proved to be highly acceptable  for various computer vision task, especially image classification \cite{8078730,8379889,SHARMA2018377,8718718}. In the proposed approach, raw fingerprint image captured by the image sensor is directly fed into customized siamese CNN.  The main idea for using CNN is that a fingerprint image has lots of global features that can easily be captured by CNN. The major factor that needs to be taken care of is the size of the image. More importantly, the image size should not shrink while going deeper into the network because a fingerprint image has regular patterns, which would be ineffective  if the image size shrinks. To keep the size of the image intact after each set of convolution layer filters performed. During the construction of the siamesse architecture, the convolution layers and the dense layers were taken into account such that minimum number of parameters are utilized in the architecture. This is done to reduce the latency of the system when deployed on the hardware. The number of the convolution layers were limited to 3 each having 4, 8 and 8 filters and batch normalisation layers respectively. At the end of the convolution layers, an average pooling is employed so that the least number of parameters are required while  passing through the dense layers. Table \ref{Table 1} presents the architecture of the siamese CNN used in the proposed CFRS.
In order to match and generate a matching score between a given pairs of fingerprint templates, a siamese network is employed \cite{7899663}. It makes use of two identical CNNs with shared weights. While training the neural network, it minimizes the distance between two similar templates and maximizes the distance between two dissimilar templates with the help of distance aware contrastive loss function . 
The contrastive loss function is defined as \cite{10.1109/CVPR.2006.100}:
\begin{equation}
  L = (1 - Y)\frac{1}{2}(D_w)^2 + (Y)\frac{1}{2}(max(0, m - D_w))^2 
\end{equation}
where
\begin{equation}
  Y = 
 \begin{cases} 
      1, & same \; class \\
      0, & different \; class\\ 
   \end{cases}
\end{equation}
\(D_w\) : Euclidean Distance between outputs vectors or embeddings of size $16\times1$ of siamese networks\\
\(m\) : margin value (dissimilar pairs beyond the margin didn't contribute to loss).\\
A schematic diagram  of the employed siamese network is presented in Fig. \ref{siamese}. The  embedding vectors of size $16 \times 1$  from the siamese network is used to compute Euclidean distance between any two templates. This Euclidean distance acts as the dissimilarity score (because CNN is trained to maximize this distance between two dissimilar templates and vice versa). Finally, the similarity score $(S_d)$ is calculated by taking the inverse of dissimilarity score.

\begin{table*} [!ht]
\caption{ CNN architecture}
\label{Table 1}
\centering
\renewcommand{\arraystretch}{1.2}
\begin{tabular}{|c|c|c|c|c|c|c|}\hline
    Layer name & input size & Filter size & No. of filters & Padding & Stride & Output size\\
    \hline
    Conv1 & $310\times240\times3$ & $3\times3$ & 4 & 1 & 1 & $310\times240\times4$\\
    \hline
    BatchNorm & $310\times240\times4$ & - & - & - & - & $310\times240\times4$\\
    \hline
    Conv2 & $310\times240\times4$ & $3\time3$ & 8 & 1 & 1 & $310\times240\times8$\\
    \hline
    BatchNorm & $310\times240\times8$ & - & - & - & - & $310\times240\times8$\\
    \hline
    Conv3 & $310\times240\times8$ & $3\time3$ & 8 & 1 & 1 & $310\times240\times8$\\
    \hline
    BatchNorm & $310\times240\times8$ & - & - & - & - & $310\times240\times8$\\
    \hline
    AveragePool & $310\times240\times8$ & $2\time2$ & - & - & 2 & $155\times120\times8$\\
    \hline
    Flatten & $155\times120\times8$ & - & - & - & - & $148800\times1$\\
    \hline
    Dense1 & $148800\times1$ & - & - & - & - & $256\times1$\\
    \hline
    Dense2 & $256\times1$ & - & - & - & - & $128\times1$\\
    \hline
    Dense3 & $128\times1$ & - & - & - & - & $16\times1$\\
    \hline
\end{tabular}

\end{table*}

\subsection{Score fusion}
As presented before, both minutiae matching module and deep learning-based module generate their respective scores represented as $S_{m}$ and $S_{d}$.
Before fusing these scores, both scores are normalized to the range $(0, 1)$ using min-max normalization \cite{minmax}. 
Finally, a weighted sum is computed as:
\begin{equation}
    S_{f} = w_{d}S_{d} + w_{m}S_{m}
\end{equation}
where $w_{d}$ and $w_{m}$ represent the weight associated with  scores obtained from deep learning and minutiae matching-based approaches. 
These weights $w_{d}$ and $w_{m}$  have been  empirically set to 0.4 and 0.6.

\subsection{ Hardware implementation}
\subsubsection{Electrical components and image capturing environment}

\begin{figure*}[h]
    \centering
    \begin{subfigure}[b]{0.6\textwidth}
    \centering
    \includegraphics[width=\textwidth]{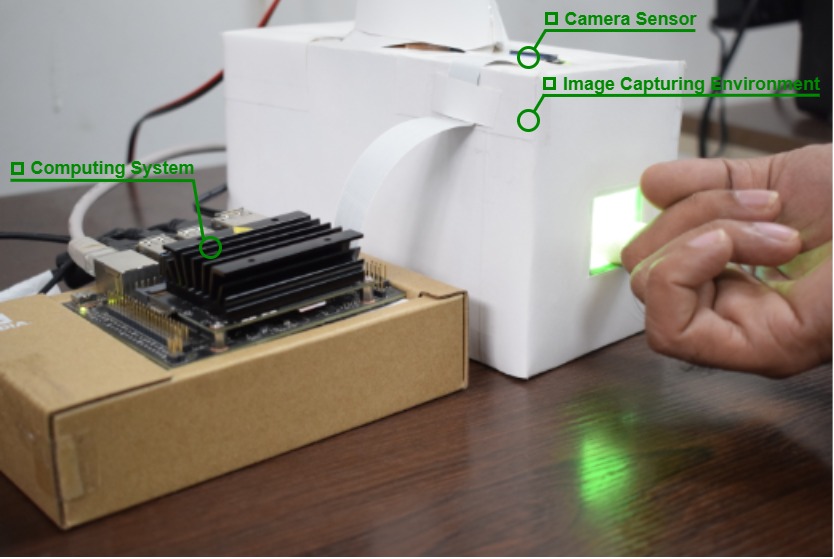}
    \caption{Image capturing environment}
    \label{capturing_environment}
    \end{subfigure}
    \hfill
    \begin{subfigure}[b]{0.29\textwidth}
    \centering
    \includegraphics[width=\textwidth]{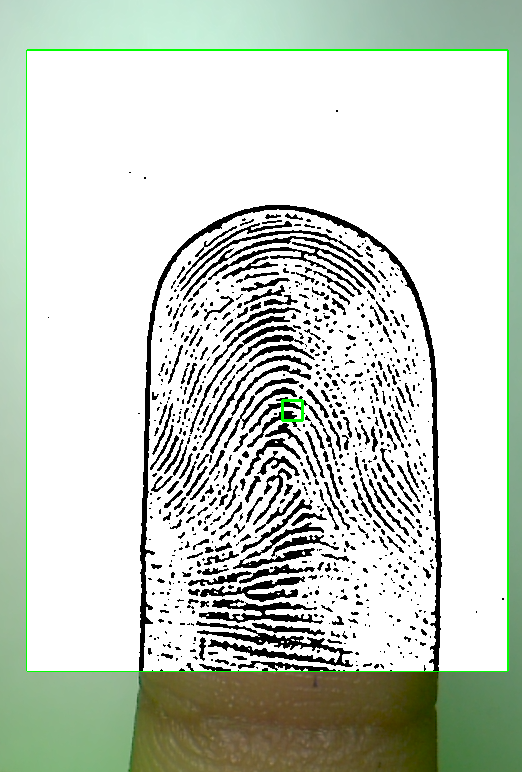}
    \caption{Real-time AMT}
    \label{distal}
    \end{subfigure}
    \caption{Pink distortion and its solution}
    \label{pink_distortion}
\end{figure*}

For our CFRS, we needed a sensor which can work in bright as well as dim light. Hence, we have used the Raspberry Pi NoIR Camera V2 \cite{rpicam}. It has a Sony IMX219 8-megapixel sensor, and it does not employ the infrared filter (NoIR = No InfraRed).
NVIDIA Jetson Nano Developer Kit \cite{jetsonnano} is used as a computing system, which is a compact, mighty embedded computer that enabled us to run our developed system. It just needs at most 5 Watts of DC power supply.
To capture the quality finger images, we designed a finger image  capturing environment (Fig. \ref{capturing_environment}) with the help of cardboard. As presented in  Fig. \ref{capturing_environment}, an LED bulb is used to illuminate the capturing environment. The camera sensor is fixed on the top of the cardboard to capture the image of the finger placed beneath it. A small square opening is made on the front of the cardboard to enter the finger inside it. Our system is designed to capture only the fingerprint on distal phalanges, on which we have applied AMT in real-time to minimize the latency (Fig. \ref{distal}).

\subsubsection{System implementation methodology}
\begin{figure}
    \centering
    \begin{subfigure}{0.21\textwidth}
    \centering
    \includegraphics[width=\textwidth]{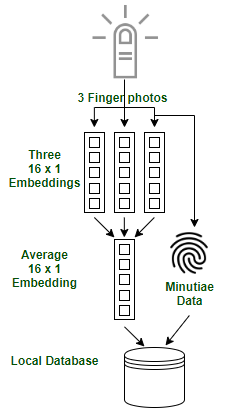}
    \caption{Implemented enrollment process}
    \label{enrollment}
    \end{subfigure}
    \hfill
    \begin{subfigure}{0.23\textwidth}
    \centering
    \includegraphics[width=\textwidth]{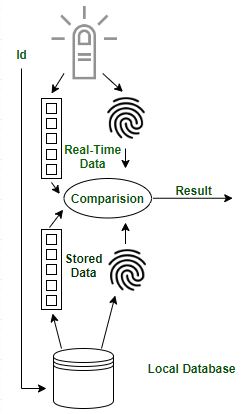}
    \caption{Implemented verification process}
    \label{verification}
    \end{subfigure}
    \caption{Enrollment and verification}
    \label{enroll_verify}
\end{figure}

The CFRS is developed to work on Verification mode. In this mode of operation, the system has two phases, namely enrollment and verification. In enrollment phase, the system captures the three finger photos of the user's finger with different placement and orientation. The average of three $16\times1$ embeddings corresponding to  three finger photos from deep-learning based approach, and minutiae data of one of the finger photo 
is stored as the template in the local database (refer to Fig. \ref{enrollment}) and a unique id is assigned to the user. In the verification phase, system first captures the finger photo of the user and asks for the unique id provided during the enrollment state. With the captured finger photo, system then generates the $16\times 1$ embedding, and minutiae data and then computes a similarity score (as discussed in score fusion) with templates of the claimed id stored in database. If the similarity score is above some threshold, the system displays a message of \textit{correct match established}  on the connected monitor screen (refer to Fig. \ref{verification}).

\subsubsection{Hardware connections}
\begin{figure}[h]
    \centering
    \includegraphics[width=0.5\textwidth]{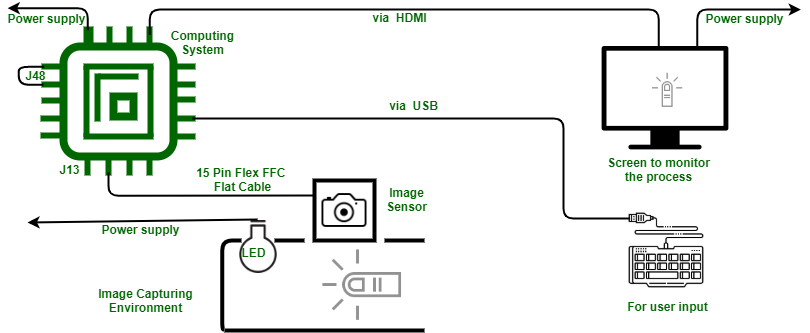}
    \caption{A schematic diagram representing connection of various components of CFRS}
    \label{connections}
\end{figure}

A detailed schematic  of all the components, along with connections, is presented in Fig. \ref{connections}.
As can be observed, the image sensor is connected to the computing system's camera connector J13 pin. A custom monitor is connected via HDMI cable along with keyboard and mouse via USB cable. The image sensor is attached to the top of the image capturing environment, which provides the flexibility to put the finger facing up inside it. All the components are powered separately.

\section{Experimental Results and Discussion}
\label{results}
\subsection{Database preparation}

As presented in Fig. \ref{capturing_environment}, we  have developed a unique contactless image capturing environment. This capturing environment is used to collect an in-house contactless fingerprint dataset referred to as IITI-CFD.  
We have captured a total of 1640 finger images from 206 fingers, each contributing eight impressions.  A sample finger image from IITI-CFD is presented in Fig. \ref{raw_image}. A detailed description of IITI-CFD is presented in Table \ref{table_2}.
For training the CNN-based approach, IITI-CFD is divided into train and test sets. Specifically, the training set consists of 840 finger images from 105 fingers.  The remaining images from 100 fingers form the test set.

\begin{table} [h]
\caption{Details of IITI-CFD}
\centering

\begin{tabular}{|>{\raggedright\arraybackslash}M{1.5cm}|M{1.5cm}|M{1cm}|M{1cm}|M{1cm}|}

\hline
\multicolumn{1}{|>{\centering\arraybackslash}M{2cm}|}{Dataset}&Image size (pixels)&Fingers&Images per finger&Total images\\
\hline
 Training set &$310\times 240$&105&8 &840\\
\hline
 Test set &$310\times 240$&100& 8 &800\\
\hline
\end{tabular}
\label{table_2}
\end{table}

\subsection{Performance measures}

For evaluating the performance of the proposed system, we have employed the following performance measures namely,  equal-error-rate (EER), FMR100 and FMR1000 \cite{eer}. Let the stored template and the verification feature set be represented as $T$ and $I$. 
For evaluating accuracy of biometrics  verification system, genuine score distribution (by comparing $T$ and $I$  from the same finger) and impostor score distribution (by comparing $T$ and $I$  from the different finger) are obtained. Based on the threshold $(th)$, there are two verification error rates namely, false match rate (FMR) and false non-match rate (FNMR). FMR is defined as the percentage of impostor pairs whose comparison score is greater than $th$ and FNMR is the percentage of genuine pairs whose comparison score is lower than $th$. 
EER denotes the error rate at $th$ for which both FMR and FNMR are identical \cite{maltoni2009handbook}. The lower the equal error rate value, the higher the accuracy of the biometric system. Generally, performance of the system is reported at all operating points ($th$) by plotting the receiver operating characteristic (ROC) curve or detection error trade-off (DET) curve. The ROC curve is a visual characterization of the trade-off between the FMR and the 1-FNMR. On the other hand,  DET curve plots the trade-off between FMR and FNMR. Specifically, the DET curve is 
 utilized to calculate FMR100 and FMR1000 of the proposed system.

\subsection{Experimental results}
The siamese network has been trained using the finger images of the training set.  Specifically, the model has been trained for 70 epochs using the adaptive moment estimation (ADAM) \cite{kingma2014method}  to optimize the loss function.
 We have employed ReLU activation function\cite{agarap2018deep} and batch normalization \cite{DBLP:journals/corr/IoffeS15}  after each convolutional layer in the proposed model. The model has been trained on the Google co-laboratory (colab) platform \cite{8485684} that uses a Tesla K80 which has 2496 CUDA cores as the GPU and a single-core hyper-threaded Xeon Processor with 2.3 GHz of processing speed. Besides, 12 GB GDDR5 VRAM was also provided by Google colab.

In order to calculate the performance metrics,  genuine scores are obtained by comparing  different impressions of the same finger and impostor scores are obtained by comparing images of different fingers. 
The feature embedding of all the test images are obtained from the output of the trained siamese network and the euclidean distances of the genuine pairs and imposter pairs are calculated and stored, which is then utilized in finding out the EER and plotting of the ROC curve with the help of FNMR and FMR by varying decision threshold. For the test dataset of 800 images (with 8 images from each finger), a total of 2800 genuine pairs and 14850 imposter pairs are obtained and the euclidean distances of their output embeddings were calculated. The genuine pairs are generated between all the possible pairs of different impressions of the same finger and is repeated for all the fingers in the test set. The Imposter pairs are generated by considering a single impression of all the fingers and finding all the possible pairs. This process is repeated for an another two sets of impressions of all the fingers in the test set. The same set of images were also used as test set for NBIS based minutiae matching.

\begin{figure}[h]
    \centering
    \includegraphics[scale=0.52]{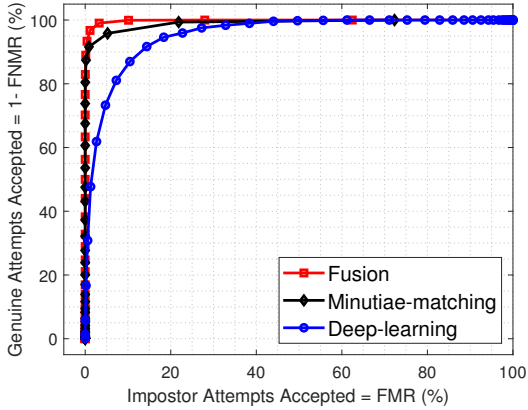}
    \caption{ROC curves for comparison between deep-learning, minutiae-matching and score-fusion technique}
    \label{ROC}
\end{figure}

The ROC curves of the proposed approach are presented in Fig. \ref{ROC}.   
As can be observed,  score-level fusion of the deep learning-based approach and minutiae-based approach provides superior results than the individual approaches. 

\begin{figure}[t]
    \centering
    \begin{subfigure}[b]{0.46\textwidth}
    \centering
    \includegraphics[width=\textwidth]{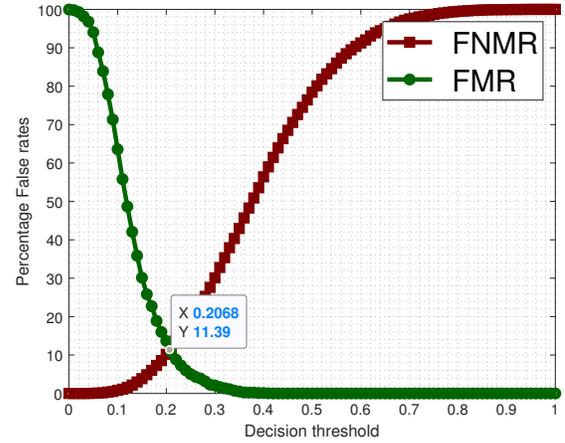}
    \caption{Deep learning based}
    \label{Dl}
    \end{subfigure}
    
    \begin{subfigure}[b]{0.46\textwidth}
    \centering
    \includegraphics[width=\textwidth]{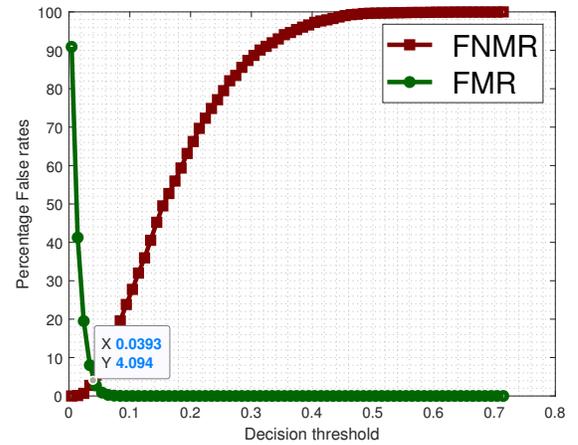}
    \caption{Minutiae matching based}
    \label{MM}
    \end{subfigure}
    
    \begin{subfigure}[b]{0.46\textwidth}
    \centering
    \includegraphics[width=\textwidth]{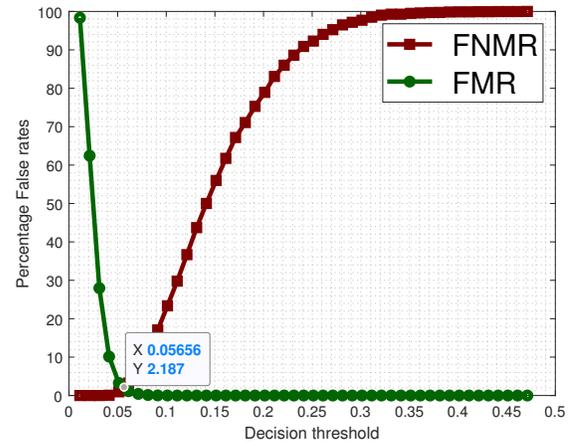}
    \caption{Score fusion based}
    \label{SF}
    \end{subfigure}
    \caption{FMR and FNMR plots}
    \label{Far_Frr}
\end{figure}

The FNMR  and FMR are plotted against varying threshold values which make the verdict for the system. The plots for the FNMR and FMR intersects at a point which intern is the minimum intersection point which gives the EER of the System. EER for the individual branches \textit{i.e}, deep learning approach,  minutiae based  approach , and score fusion approach are calculated from FMR vs FNMR plots (refer to Fig. \ref{Far_Frr}) and are summarized in Table. \ref{table_3}.

\begin{figure}[h]
    \centering
    \includegraphics[width = 0.5 \textwidth]{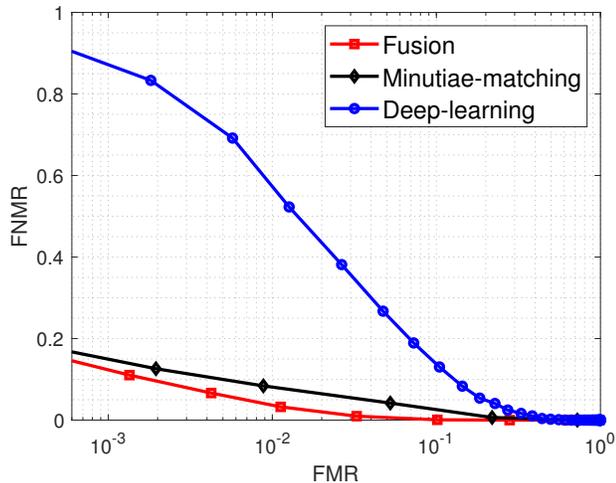}
    \caption{DET Curve}
    \label{DET}
\end{figure}

\begin{table} [h]
\caption{ EER comparison}
\label{EER}
\centering

\begin{tabular}{|>{\raggedright\arraybackslash}M{3.5cm}|M{3.5cm}|}

\hline
\multicolumn{1}{|>{\centering\arraybackslash}M{2cm}|}{Approach}&EER\\
\hline
 Deep Learning & $11.39\%$\\
\hline
 Minutiae Matching & $4.09\%$\\
\hline
Score Fusion & $2.19$\%\\
\hline
\end{tabular}
\label{table_3}
\end{table}

Further,  DET curve depicts the error-rate trade-off at all possible operating points has been used to obtain FMR$100$ and FMR$1000$. Fig. \ref{DET} presents DET curves   of three approaches. The FMR$100$ and FMR$1000$ obtained from DET curve are summarized in Table. \ref{table_4}.
\begin{table} [h]
\caption{FMR100 and FMR1000}
\centering

\begin{tabular}{|>{\raggedright\arraybackslash}M{2.5cm}|M{2.5cm}|M{2.5cm}|}

\hline
\multicolumn{1}{|>{\centering\arraybackslash}M{2cm}|}{Approach}& FMR$100$ & FMR$1000$ \\
\hline
 Deep Learning & $0.573$ & $0.872$\\
\hline
 Minutiae Matching & $0.081$ & $0.150$\\
\hline
Score Fusion & $0.037$ & $0.123$\\
\hline
\end{tabular}
\label{table_4}
\end{table}
As can be observed from Table \ref{EER} and Table \ref{table_4},  EER, FMR100 and  FMR1000 which have been calculated for the test dataset in all the three approaches suggest that the fusion of the deep learning score with the state-of-the-art NBIS software gives us the best results as compared to the methods taken individually. The fused score value is then compared with a threshold which gives us the final verdict of the developed biometric system.

\section{Conclusion}
In this work, we have identified main issues with the contact-based biometric system and unfolded the scope of the contact-less biometric system. Apart from standard image processing and state-of-art feature extraction algorithms, deep learning models can be used to improve matching accuracy to even better. With the advancement in sensing technology and computation power, the contact-less domain has an enormous market scope.
Our results show that contact-less biometric systems can also achieve similar accuracy, which other systems claim in the commercial market.

As part of future work, we would like to implement our developed model on a micro-controller and GPU for faster computation and embed it on the Printed Circuit Board (PCB) along with image sensor and image capturing environment. This way, our developed model would have the potential of the standalone embedded device.


%





\ifCLASSOPTIONcaptionsoff
  \newpage
\fi



%


\bibliographystyle{IEEEtran}
\bibliography{IEEEabrv,BTP}

%








\end{document}